\pdfoutput=1

\documentclass[11pt]{article}

\usepackage{emnlp2021}

\usepackage[T1]{fontenc}

\usepackage[utf8]{inputenc}

\usepackage{microtype}

\usepackage{times}
\usepackage{latexsym}
\usepackage{hyperref}
\usepackage{graphicx}
\usepackage{graphics}
\usepackage{multirow}
\usepackage{amsmath}
\usepackage{amsfonts}
\usepackage{amssymb}
\usepackage{tablefootnote}
\usepackage{color}
\usepackage{bbm}
\usepackage{graphicx}
\usepackage{xspace}
\usepackage{tikz}
\usepackage{pgfplots}
\usepackage[normalem]{ulem}
\usepackage{xcolor}
\usepackage{dashbox}
\usepackage{textcomp}
\usepackage{tcolorbox}
\usepackage{adjustbox}
\usepackage{url}
\usepackage{subcaption}
\usepackage{fdsymbol}

\pgfplotsset{compat=1.13}
\usetikzlibrary{calc,fit,positioning,arrows}
\tikzstyle{line}=[draw]
\pgfdeclarelayer{bg}
\pgfsetlayers{bg,main}

\tikzset{
	keep name/.style={
		prefix after command={
			\pgfextra{\let\fixname\tikzlastnode}
		}
	},
	partialbox/.style={
		keep name,
		append after command={
			(\fixname.north) -- 
			(\fixname.north west) -- 
			(\fixname.south west) -- 
			([xshift=-#1]\fixname.south)
			(\fixname.north) -- 
			(\fixname.north east) -- 
			(\fixname.south east) -- 
			([xshift=#1]\fixname.south)
		}
	},
	partialbox/.default=5pt
} 

\tikzstyle{vecArrow} = [thick, decoration={markings,mark=at position
   1 with {\arrow[semithick]{open triangle 60}}},
   double distance=1.4pt, shorten >= 5.5pt,
   preaction = {decorate},
   postaction = {draw,line width=1.4pt, white,shorten >= 4.5pt}]
\tikzstyle{innerWhite} = [semithick, NavyBlue,line width=1.4pt, shorten >= 4.5pt]

\newcommand{\La}{\mathcal{L}}
\newcommand{\model}{{\fontfamily{lmtt}\selectfont ADAPET}\xspace}
\newcommand{\PET}{\textsc{Pet}}

\newcommand{\fancyp}{\textbf{\fontfamily{lmtt}\selectfont p}}
\newcommand{\fancyq}{\textbf{\fontfamily{lmtt}\selectfont q}}
\newcommand{\fancyh}{\textbf{\fontfamily{lmtt}\selectfont h}}
\newcommand{\fancye}{\textbf{\fontfamily{lmtt}\selectfont e}}
\newcommand{\fancyw}{\textbf{\fontfamily{lmtt}\selectfont w}}
\newcommand{\fancyci}{\textbf{\fontfamily{lmtt}\selectfont $\text{c}_\text{i}$}}
\newcommand{\fancysi}{\textbf{\fontfamily{lmtt}\selectfont $\text{s}_\text{i}$}}
\newcommand{\textblue}[1]{\textbf{\textcolor{NavyBlue}{#1}}}

\newcommand{\reddiamond}{\textcolor{red}{ \vardiamondsuit}}

\definecolor{c0}{cmyk}{1,0.3968,0,0.2588} 
\definecolor{c1}{cmyk}{0,0.6175,0.8848,0.1490} 
\definecolor{c2}{cmyk}{0.1127,0.6690,0,0.4431} 
\definecolor{c3}{cmyk}{0.6765,0.2017,0,0.0667} 
\definecolor{c4}{cmyk}{0.3081,0,0.7209,0.3255} 
\definecolor{c5}{cmyk}{0,0.8765,0.7099,0.3647}
\definecolor{darkgrey}{RGB}{180,180,180}
\definecolor{decentgrey}{RGB}{220,220,220}

\newtcbox{\pattern}{on line,colback=gray!30,colframe=black,size=fbox,arc=3pt, box align=base,before upper=\strut,
top=-2pt, bottom=-2pt, boxrule=0pt}
\newtcolorbox{multipattern}{on line,colback=gray!30,colframe=black,size=fbox,arc=3pt, box align=base, top=-2pt, bottom=0pt, boxrule=0pt, before=\adjustbox{}\bgroup, after=\egroup, before upper=\strut}

\title{Improving and Simplifying Pattern Exploiting Training}

\author{Derek Tam\thanks{\hspace{0.5em}Equal contribution} \and Rakesh R Menon\footnotemark[1] \AND Mohit Bansal \and Shashank Srivastava \and Colin Raffel\\
    UNC Chapel Hill \\ 
  \texttt{\{dtredsox, rrmenon, mbansal, ssrivastava, craffel\}@cs.unc.edu}
}

\begin{document}
\maketitle
\begin{abstract}
Recently, pre-trained language models (LMs) have achieved strong performance when fine-tuned on difficult benchmarks like SuperGLUE. However, performance can suffer when there are very few labeled examples available for fine-tuning. Pattern Exploiting Training (\PET) is a recent approach that leverages patterns for few-shot learning. However, {\PET} uses task-specific unlabeled data. In this paper, we focus on few shot learning without any unlabeled data and introduce \model, which modifies \PET's objective to provide denser supervision during fine-tuning. As a result, \model outperforms {\PET} on SuperGLUE without any task-specific unlabeled data. Our code can be found at \url{https://github.com/rrmenon10/ADAPET}. 

\end{abstract}

\section{Introduction}

Pre-trained language models (LMs) have shown significant gains across a wide variety of natural language processing (NLP) tasks in recent years \cite{devlin2018bert, radford2018improving, raffel2020exploring}. Most of these gains are obtained by fine-tuning language models on labeled data for a particular task. However, performance can suffer when there is very limited labeled data available for a downstream task \cite{xie2019uda, chen2020mixtext}. 

Recently, GPT-3 \cite{brown2020language} demonstrated how language models, when scaled to hundreds of billions of parameters, can learn well when primed with only a few labeled examples. However, the scale of GPT-3 (175B parameters) makes it impractical to study. There is, therefore, a need to develop smaller language models that can work equally well with limited labeled data.

\begin{figure}
    \centering
	\begin{tikzpicture}
	\begin{axis}[
	cycle list name=color list,
	xlabel={\sffamily\small Amount of task-specific data used},
	ylabel={\sffamily\small SuperGLUE Performance},
	axis line style={decentgrey!95!black},
	grid=major,
	major grid style={line width=.2pt,draw=decentgrey},
	ymin = 70,
	ymax = 78,
	xmin = 10,
	xmax = 100000,
	xmode = log,
	minor tick style={decentgrey!0},
	major tick style={decentgrey},
	log basis x={10},
	xtick pos=left,
	ytick pos=left,
	ylabel near ticks,
	xlabel near ticks,
	xticklabels={$10$, $10^2$, $10^3$, $10^4$, $10^5$},
	tick align=outside,
	tick label style={font=\footnotesize},
	major tick length=0.075cm,
	width = \linewidth,
	height = 0.25\textheight,
	log ticks with fixed point,
	x tick label style={/pgf/number format/1000 sep=\,},
	]
	\addplot[mark=*, c0, thick, mark options={solid}] coordinates {
		(32,76.0) 
	} node[right,pos=1,xshift=0.025cm](adapet){\model};
	
	\addplot[mark=*, c3, thick, mark options={solid}] coordinates {
		(32,71.8) 
	} node[right,pos=1,xshift=0.025cm](gpt){\sffamily GPT-3};
	
	\addplot[mark=*, c1, thick, mark options={solid}] coordinates {
		(9310.375,74.0)
	} node[right,pos=1,xshift=0.025cm](PET){\sffamily PET};
	
	\addplot[mark=*, c2, thick, mark options={solid}] coordinates {
		(9310.375,75.4)
	} node[right,pos=1,xshift=0.025cm](iPET){\sffamily iPET};

	\path[line] (axis cs:6000,76.7) -- node[above=0.2em]{\textbf{$0.3\%$ of the data}} (axis cs:100, 76.7); 
	\draw[thick, ->] (axis cs:6000,76.7) -- (axis cs:100, 76.7);

	\end{axis}
	\end{tikzpicture}
	\caption{Performance of \model vs i{\PET}/{\PET}
          and GPT-3 on SuperGLUE. While i{\PET}/{\PET} are parameter-efficient, they use $\sim$9K unlabeled examples in addition to 32 labeled examples per task. \model uses just 32 labeled examples, and performs better than i{\PET}.} 
	\label{figure:intro}
\end{figure}
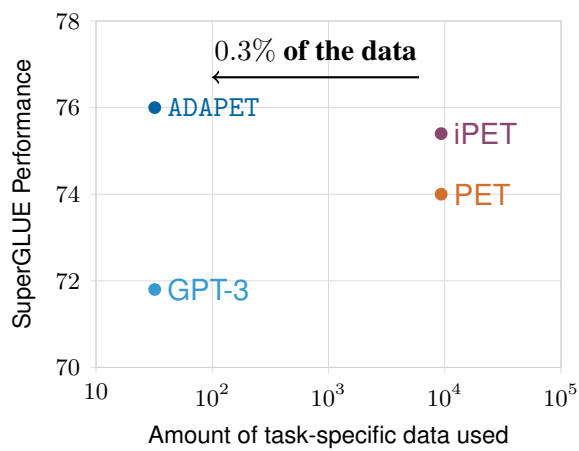 

Pattern-Exploiting Training \citep[\PET;][]{schick2020exploiting, schick2020s}  reformulates natural language understanding tasks as cloze-style questions and performs gradient-based fine-tuning. In doing so, {\PET} outperforms GPT-3 with few labeled examples using ALBERT \cite{lan2019albert}. However, {\PET} uses additional task-specific unlabeled data.

We propose \model (\textbf{A} \textbf{D}ensely-supervised \textbf{A}pproach to \textbf{P}attern \textbf{E}xploiting \textbf{T}raining) that uses more supervision by decoupling the losses for the label tokens and a label-conditioned masked language modeling (MLM) objective over the full original input. On SuperGLUE \cite{wang2019superglue} with 32 labeled examples per task, \model outperforms i{\PET} without any unlabeled data. 

\begin{figure*}[t!]
   \vspace{3mm}
    \centering
    \begin{subfigure}[t]{0.5\textwidth}
        \centering
        \includegraphics[height=1.6in]{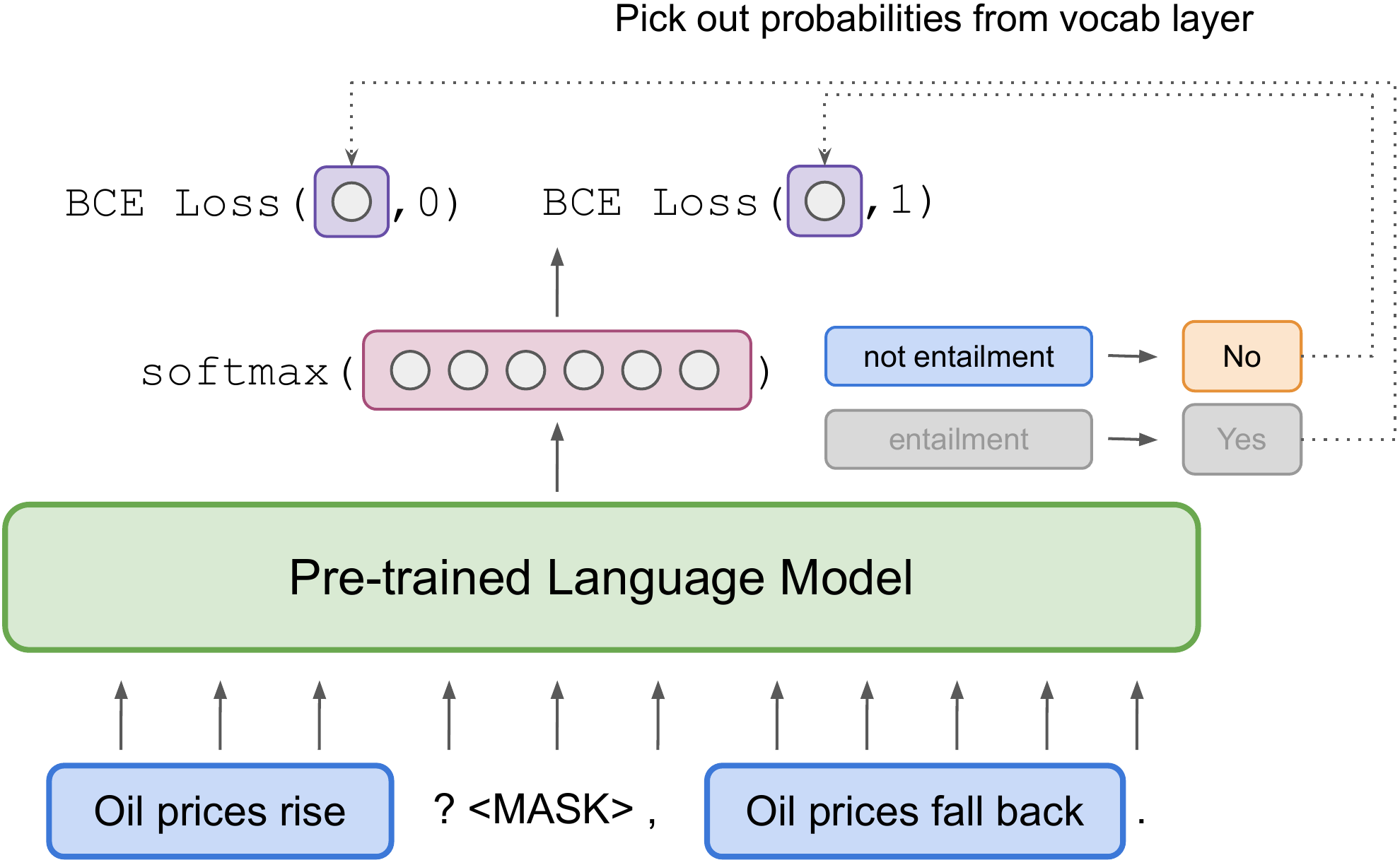}
        \caption{Decoupling Label Loss}
        \label{fig:decouple}
    \end{subfigure}%
    \begin{subfigure}[t]{0.5\textwidth}
        \centering
        \includegraphics[height=1.6in]{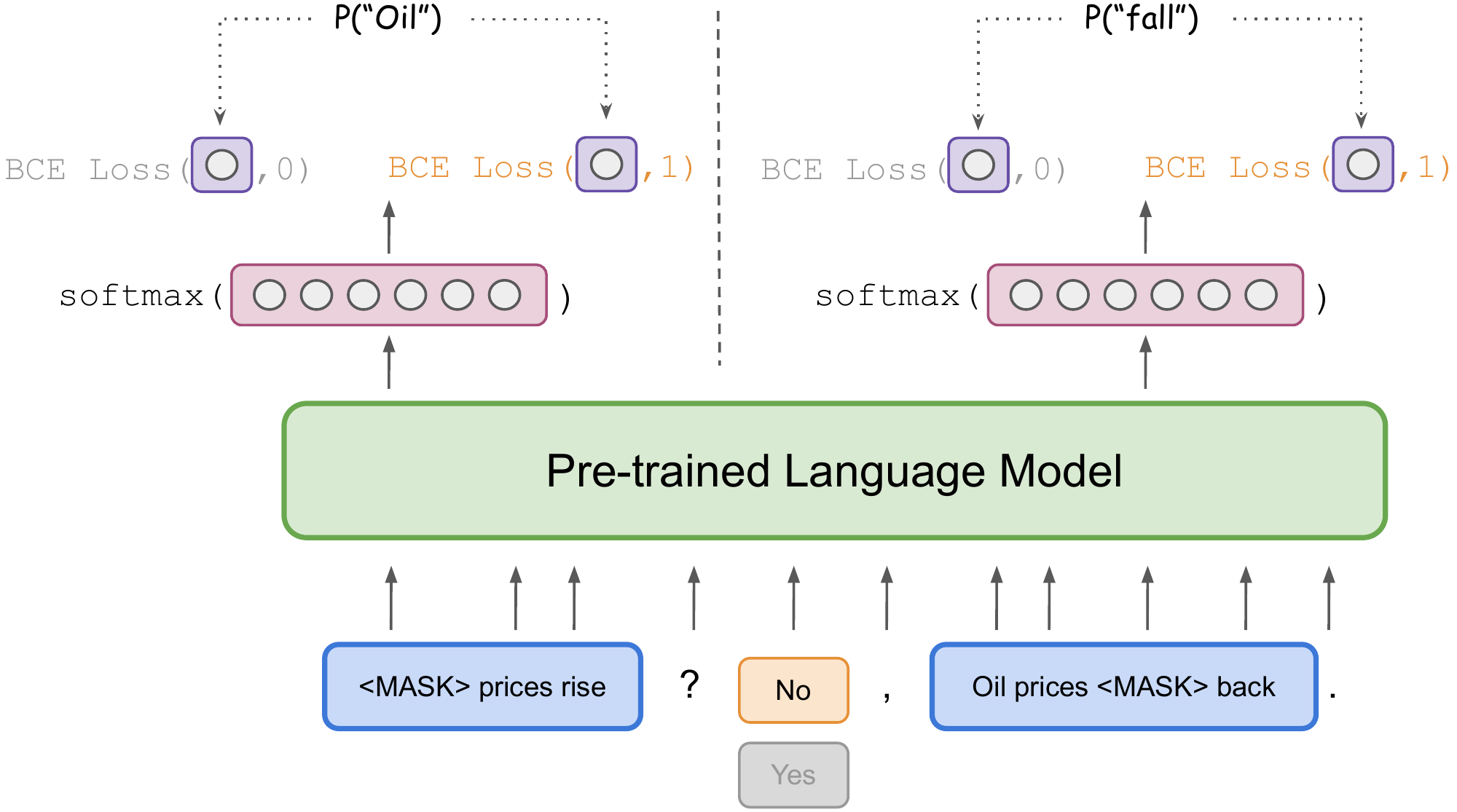}
        \caption{Label Conditioning}
        \label{fig:mlm}
    \end{subfigure}
    \caption{We illustrate the training with the two components of \model. Here, the \textblue{blue} boxes refer to the inputs from a task (entailment, in this case). Figure \ref{fig:decouple} shows the decoupling label objective. The model has to predict the correct and incorrect labels at the masked out position, using a BCE loss over all labels. For the label conditioning objective in Figure \ref{fig:mlm}, the input text either includes the correct or incorrect label. At a randomly masked out position, the model should predict the original token when the input text has the correct label, and should not predict the original token when the input text has an incorrect label.}
\end{figure*}

\section{Background} \label{sec:bkg}

\textbf{Cloze-style questions and MLM.} A cloze task is a problem where certain parts of a text are removed, and the goal is to replace the missing portion based on the context \cite{taylor1953cloze}. Here, the text that has some parts removed is considered a cloze-style question. Inspired by cloze tasks, BERT introduces the MLM objective that tries to predict the original word at the masked out positions in a cloze question.\\
\\
\textbf{Notation.} Let $G$ represent a language model, $x$ represent the input example converted into a cloze-style question, and $y$ represent the label at the masked location $m$. We are interested in the quantity $[\![G_m(x)]\!]_{z}$ which represents the logit value for a specific token $z$ at the mask location $m$.

\subsection{Unlabeled Data Access} \label{sec:few_shot_def}

 \citet{schick2020exploiting,schick2020s} assumes access to task-specific unlabeled data. For some applications such as sentiment analysis, unlabeled data can be cheap to acquire. But for SuperGLUE, where the examples are pairs of text with a label that is constructed to test a model's natural language understanding abilities, it might be more expensive to acquire unlabeled data. For example, the construction of BoolQ requires annotators to filter good question-article pairs before assigning labels \cite{clark2019boolq}. Hence, for our setup, we do not assume access to task-specific unlabeled data, which aligns with the setup in \citet{brown2020language}. 

\subsection{\PET} \label{sec:pet}

Our work primarily builds on top of {\PET} \cite{schick2020exploiting, schick2020s}. {\PET} converts an example into a cloze-style question, similar to the input format used during pre-training. The query-form in {\PET} is defined by a Pattern-Verbalizer Pair (PVP). Each PVP consists of 
\begin{itemize}
    \item a \textbf{pattern} which describes how to convert the inputs into a cloze-style question with masked out tokens. We illustrate this for an entailment task in Figure \ref{fig:decouple}. Here, we convert the premise (``\textit{Oil prices fall back}") and the hypothesis (``\textit{Oil prices rise}") into a cloze-style question with the pattern: $<$\textit{premise}$>$ ? $<$\textit{mask}$>$, $<$\textit{hypothesis}$>$.
    \item a \textbf{verbalizer} which describes the way to convert the classes into the output space of tokens. In Figure \ref{fig:decouple}, the verbalizer maps \textit{``Not Entailment/Entailment"} to  \textit{``No/Yes''}.
\end{itemize}
After hand-designing a PVP for a given task, {\PET} obtains logits from the model $G_m(x)$ (in the single-token label case). Given the space of output tokens $\mathcal{Y}$, (in Figure \ref{fig:decouple} $\{``\text{\textit{Yes}}", ``\text{\textit{No}}"\}$) {\PET} computes a softmax over $y \in \mathcal{Y}$, using the logits from $G_m(x)$. The final loss is shown in Equation \ref{eq:pet}. 

{\small
\begin{align}
    q(y|x) &= \frac{\exp([\![G_m(x)]\!]_{y})}{\sum\limits_{y' \in \mathcal{Y}} \exp([\![G_m(x)]\!]_{y'})} \\
    \La &= \texttt{CE} (q(y^*|x), y^*) \ 
    \label{eq:pet}
\end{align}
}%
{\PET} additionally distils knowledge from an ensemble of models trained with different patterns on both labeled and unlabeled data. i{\PET} is an iterative variant of {\PET} that trains models across iterations. The size of the training set gradually increases each iteration based on the labels of previous iterations.  
For a description of the different patterns used across the tasks \cite{schick2020s}, we refer the reader to Appendix \ref{sec:app_pvp}.

\section{\model} 

Our proposed approach, called \model, modifies the objective from {\PET} so that it can provide more supervision and learn without task-specific unlabeled data.

\subsection{Decoupling Label Losses}

{\PET} computes class probabilities using the logits that correspond to the labels for a specific task. This discards the information from all the other logits in the vocabulary that do not correspond to a label.  For example, in Figure \ref{fig:decouple}, \textit{``oil"} is not a class token so the LM head should assign a low probability to \textit{``oil"}. However, because {\PET} only extracts the token logits that correspond to labels, the non-label tokens will never have any gradient signal. 

One solution is to change the objective to a regular MLM objective. In that case, there would be no distinction between tokens corresponding to incorrect classes and any other token in the vocabulary. For example, in Figure \ref{fig:decouple}, the model would be trained to treat \textit{``Yes"} (the incorrect token) the same as any other token such as \textit{``oil"}. While we want the model to discourage \textit{``oil"}, the training objective should still specifically suppress \textit{``Yes"}. 

In \model, we penalize incorrect class tokens and encourage correct class tokens. Specifically, the model computes the probability of each token as a softmax normalized across all tokens so that each probability is influenced by the logits of all the vocabulary tokens. Then, we maximize the probability of the correct class tokens and minimize the probability of incorrect class tokens. This is equivalent to binary cross entropy, as shown in Figure \ref{fig:decouple}. 
Formally, if $y^*$ is the true label for an example,

{\small
\begin{align}
    q(y|x) &= \frac{\exp([\![G_m(x)]\!]_{y})}{\sum\limits_{v' \in \mathcal{V}} \exp([\![G_m(x)]\!]_{v'})} \\
    \La_D &= \log q(y^*|x) - \sum\limits_{y \neq y^*}  \log q(y|x)
\end{align}
}%

The loss can be rewritten using binary cross entropy or regular cross entropy as:
{\small
\begin{align}
    \label{eq:disc_1}
    \La_D &= \texttt{BCE} (q(y^*|x), 1) + \sum\limits_{y \neq y^*} \texttt{BCE}(q(y|x), 0)\\ 
    \label{eq:disc_2}
    &= \texttt{CE}(q(y^*|x), y^*) - \sum\limits_{y \neq y^*} \texttt{CE}(q(y|x), y) 
\end{align}
}%

\subsubsection{Unified Loss for Different Tasks}

For normal tasks where the label is exactly one token, {\PET} uses the formulation described in Equation \ref{eq:pet}. For WSC \cite{levesque2012winograd}, which does not have incorrect class labels, {\PET} uses the original MLM objective rather than Equation \ref{eq:pet}. This is equivalent to Equation \ref{eq:disc_1} without the second term in \model.

For other tasks with multi-token labels (COPA \cite{roemmele2011choice}, ReCoRD \cite{zhang2018record}), {\PET} computes the probability of the classes as the sum of the log probabilities of the individual tokens. However, it is not obvious how to convert these label probabilities into a valid probability distribution.

Rather than normalizing the probabilities, {\PET} uses a hinge loss to ensure a margin between the correct label and the incorrect labels.

In \model, for each token in the label, $\mathcal{L}_D$ discriminates the correct token from every other tokens, via the following loss:\footnote{We ignore tokens that are common in all labels.}

{\small 
\begin{align}
    \label{eq:disc_multi}
    \La_D &= \hspace{-0.5em} \sum\limits_{z^{*} \in y^{*}} \hspace{-0.6em}\texttt{BCE} (q(z^{*}|x), 1) + \sum\limits_{y \neq y^*} \sum\limits_{z \in y}\texttt{BCE}(q(z|x), 0)
\end{align}
}%
 
This objective splits a \textit{single loss based on multiple tokens into multiple losses over single tokens}. As a result, we do not need to to multiply the probabilities of the individual tokens, and thus do not run into normalization issues. 

\begin{table*}[ht!]
    \centering
    \resizebox{1\textwidth}{!}{%
    \begin{tabular}{l l l l l l l l l | l}
        \hline 
        & \textbf{BoolQ} & \textbf{CB} & \textbf{COPA} & \textbf{RTE} & \textbf{WiC} & \textbf{WSC} & \textbf{MultiRC} & \textbf{ReCoRD} & \textbf{Avg} \\
        \textbf{Method} & Acc. & Acc./F1 & Acc. & Acc. & Acc. & Acc. & EM/F1a & Acc./F1 & -\\
         \hline
         ALBERT   & 55.7 & 68.6 / 49.1& 63.0 & 50.5& 41.4 & 81.7 & 3.6 / 49.8& 84.1/83.5 & 57.7\\
         \hline
         GPT-3 \textsc{(lab; single)}   & 77.5 & 82.1 / 57.2 & 92.0 $\clubsuit$ & 72.9 & 55.3 $\clubsuit \reddiamond$ & 75.0 & 32.5 / 74.8 & 89.0 / 90.1 $\clubsuit \reddiamond$ & 73.2 \\
         s{\PET} \textsc{(lab; single)}  & 76.9 &  87.5 / 85.4 & 89.0 & 67.1 & 49.7 & 82.7 $\clubsuit \reddiamond$ & 31.2 / 74.6 & 85.0 / 91.9 & 74.2 \\
        \model \textsc{(lab; single)}   & 80.3 $\clubsuit$ & 89.3 / 86.8 $\clubsuit$ & 89.0 & 76.5 $\clubsuit \reddiamond$ & 54.4 & 81.7 & 39.2 / 80.1 $\clubsuit \reddiamond$ & 85.4 / 92.1  &  77.3 $\clubsuit \reddiamond$ \\
         \hline
         {\PET} \textsc{(lab + unlab; ensemble)} & 79.4& 85.1 / 59.4& 95.0 $\reddiamond$ & 69.8 & 52.4 & 80.1 & 37.9 / 77.3 & 86.0 / 86.5  & 74.1 \\
         i{\PET}  \textsc{(lab + unlab; ensemble)} & 80.6 $\reddiamond$ & 92.9 / 92.4 $\reddiamond$ & 95.0 $\reddiamond$ & 74.0 & 52.2& 80.1 & 33.0 / 74.0 &  86.0 / 86.5 & 76.8 \\ 
        \hline
    \end{tabular}
    }
    \caption{Few-shot classification results on SuperGLUE with 32 labeled examples on the dev set. Note, we do not have access to the train split of GPT-3, so we follow the split provided by \cite{schick2020s}. \textsc{$\clubsuit$=best single pattern model, $\reddiamond$=best model overall, lab=labeled data, unlab=unlabeled data}}
    \label{tab:res}
\end{table*}

\begin{table*}[ht!]
    \centering
    \resizebox{1\textwidth}{!}{%
    \begin{tabular}{l l l l l l l l l | l}
        \hline 
        & \textbf{BoolQ} & \textbf{CB} & \textbf{COPA} & \textbf{RTE} & \textbf{WiC} & \textbf{WSC} & \textbf{MultiRC} & \textbf{ReCoRD} & \textbf{Avg} \\
        \textbf{Method} & Acc. & Acc./F1 & Acc. & Acc. & Acc. & Acc. & EM/F1a & Acc./F1 & -\\
         \hline
         GPT-3 \textsc{(lab; single)}  & 76.4 & 75.6 / 52.0 & 92.0 $\clubsuit \reddiamond$ & 69.0 & 49.4 & 80.1 & 30.5 / 75.4 & 90.2 / 91.1 $\clubsuit \reddiamond$ & 71.8\\
        \model \textsc{(lab; single)}  & 80.0 $\clubsuit$ & 92.0 / 82.3 $\clubsuit \reddiamond$ & 85.4 & 75.0 $\clubsuit \reddiamond$ & 53.5 $\clubsuit \reddiamond$ &  85.6 $\clubsuit$  & 35.7 / 76.2 $\clubsuit$ & 85.5 / 86.1 & 76.0 $\clubsuit \reddiamond$\\
        \hline
         {\PET} \textsc{(lab + unlab; ensemble)} & 79.1 & 87.2 / 60.2 & 90.8 & 67.2 & 50.7  & 88.4 $\reddiamond$ & 36.4 / 76.6 $\reddiamond$ & 85.4 / 85.9 & 74.0\\
         i{\PET}  \textsc{(lab + unlab; ensemble)} &  81.2 $\reddiamond$ & 88.8 / 79.9 & 90.8 & 70.8 & 49.3 & 88.4  $\reddiamond$ & 31.7 / 74.1 & 85.4 / 85.9 & 75.4   \\ 
        \hline
    \end{tabular}
    }
    \caption{Few-shot classification results on SuperGLUE with 32 labeled examples on the hidden test set. \textsc{$\clubsuit$=best single pattern model, $\reddiamond$=best model overall, lab=labeled data, unlab=unlabeled data}} 
    \label{tab:res_test}
\end{table*}

\subsection{Label Conditioning} \label{sec:lab_cond}

The {\PET} objective encapsulates the question: "\textit{Given the input, what is the right label?}." However, since the input space and output space both consist of tokens, we can also ask the inverse question, \textit{``Given the answer, what is the correct context?"}. The model is trained to predict the input given the label. Formally, let $x'$ be the original input $x$ modified by randomly masking out tokens from the context and $x^m$ be the original context tokens masked out in $x'$. In the label conditioning objective, we are interested in the quantity $P(x^m | x', y)$, which encourages the model to predict the masked out tokens in the input given the label.

During training, if the label is correct, the model has to predict the original token, as shown in 
Figure \ref{fig:mlm}. Additionally, if the label is wrong, the model is forced to \textit{not} predict the original token.
\footnote{This assumes the context only makes sense with the correct label. Empirically though, we find this to be reasonable.}
We maximize $P(x^m|x', y^*)$ and minimize $P(x^m | x', y) \hspace{1mm}  \forall \hspace{1mm} y \neq y^*$. This objective is the same as the decoupling label losses approach described in Equation \ref{eq:disc_1}, except with different inputs and outputs.

{\small
\begin{equation}
\label{eq:mlm_eq}
    q(x^m|x', y) = \frac{\exp([\![G_m(x', y)]\!]_{x^m})}{\sum\limits_{v' \in \mathcal{V}} \exp([\![G_m(x', y)]\!]_{v'})} 
\end{equation}
}%

{\small
\begin{align}
    \La_M  = \hspace{-0.2em} \texttt{BCE} (q(x^m|x', y^{*}), 1) + \hspace{-0.5em} \sum\limits_{ y \neq y^*} \hspace{-0.4em}\texttt{BCE}(q(x^m|x', y), 0)  
\end{align}
}%
The final loss for  \model is a sum of the decoupled label loss and the label-conditioned MLM loss.

\section{Results and Analyses} \label{sec:exp}

We run experiments on SuperGLUE, and follow the same data split as \citet{schick2020s}, which consists of 32 labeled examples for each task. 

Our code is implemented in Pytorch \cite{paszke2019pytorch} using HuggingFace \cite{wolf2019huggingface}. We use the same pre-trained model and hyperparameters as {\PET}, except we increased the number of training batches to 1k and choose the best checkpoint on the dev set, since it has been shown that training longer can help even with few samples \cite{zhang2021revisiting}. For all ablation experiments, we only use the first pattern\footnote{The first pattern for each task can be found in App. \ref{sec:app_pvp}} and train for 250 batches. We refer the reader to Appendix \ref{sec:exp_det} for more details.

Since we do not assume access to unlabeled data (see Section \ref{sec:few_shot_def}), we do not apply the three-step training procedure of {\PET} and i{\PET} to \model. We still assume access to the full development set to choose the best masking ratio and checkpoint model, since {\PET} presumably used the full development set to choose their hyperparameters which we copy.  

\subsection{Results}

Table \ref{tab:res} and Table \ref{tab:res_test} shows our results on the validation and test sets on SuperGLUE. We compare against GPT-3 and {\PET}/i{\PET}. Note that {\PET}/i{\PET} uses unlabeled data and a three step training procedure \cite{schick2020s}. For fair comparison, we train {\PET} with a single pattern (s{\PET}) for 1k batches, and report scores for the best performing pattern on the validation set. We include a further analysis of how well the models perform for each pattern in Appendix \ref{sec:ind_pvp}.  

On the dev set, \model outperforms all models that do not use unlabeled data, and even outperforms PET's iterative variant, i{\PET}, by $0.5$ points absolute. Surprisingly, s{\PET} outperforms {\PET}, but still loses to i{\PET} by $2.6$ points. But, this is in line with the ablation from \citet{schick2020s}, which shows that ensembling s{\PET} models, trained with only labeled data, outperforms {\PET}. Also, \citet{gao2020making} show that the model with the best performing pattern outperforms ensembling s{\PET} models.

On the test set, \model outperforms all other models including i{\PET} without access to the unlabeled examples ($\sim$9k on average per task) and achieves state-of-the-art for few-shot learning on SuperGLUE.  
\subsection{Loss Ablation}

Table \ref{tab:ablation} shows our ablation analysis for the loss functions we introduce in this paper. From the results, we see that label conditioning (\textsc{LC}) is extremely beneficial for \model, especially on CB. Comparing our modified decoupled label objective (\model \textsc{w/o LC})  with s{\PET}, we see that it does worse for CB on F1, but does much better on RTE and MultiRC. Next, we compare against LC conditioned only on the correct label. We see that this hurts on BoolQ, but helps on CB. We ablate other model choices in Appendix \ref{sec:ablat}.

\begin{table}[ht!]
    \centering
    \resizebox{0.5\textwidth}{!}{%
    \begin{tabular}{l c c c c c}
        \hline 
        &  \textbf{BoolQ} & \textbf{CB} & \textbf{RTE} & \textbf{MultiRC} \\
        \textbf{Method} &  Acc. & Acc./F1 &  Acc. & EM / F1a\\
         \hline
        \model  &  \textbf{79.4} & \textbf{91.1 / 88.1} & \textbf{75.1} & \textbf{38.6 / 79.8} \\
        \model \small{\textsc{w/o LC}}  &  78.1 & 75.0 / 62.8  & 64.3 & 37.0 / 79.1 \\
        \model \small{\textsc{LC (Pos. Ex. only)}}  &  75.4 &	83.9 / 80.9  & 72.2 & 31.3 / 76.9 \\
        \hline
         s{\PET}  & 77.5&  75.0 / 72.8 & 57.0 & 26.5 / 73.2 \\
         \hline
    \end{tabular}
    }
    \caption{Ablation of \model with different components. Best numbers have been \textbf{bolded}. (\textsc{LC= Label Conditioning})}
    \label{tab:ablation}
\end{table}

\section{Conclusion}
In this paper, we propose \model, a new method for few-shot natural language understanding. Crucially, our work does not use unlabeled data and instead leverages more supervision to train the model. Assuming the same data budget, our model outperforms GPT-3 on SuperGLUE using just $0.1\%$ as many parameters. However, our method has limitations; for example, we use a naive random masking strategy, which might not make sense for label conditioning. Future work could look into better masking strategies for labeled conditioned MLM, such as masking important tokens based on the the gradients of the logits for an example, as has been done for interpreting models \cite{simonyan2014saliency}. 

\section*{Acknowledgments}
This work was supported in part by ONR Grant N00014-18-1-2871, DARPA YFA17-D17AP00022, and NSF-CAREER Award 1846185. The views contained in this article are those of the authors and not of the funding agency.

\bibliography{emnlp2021}
\bibliographystyle{acl_natbib}

\appendix

\section*{Appendix}

\section{Patterns and Pattern Performances}
\subsection{Pattern Verbalizer Pairs}  \label{sec:app_pvp}
We list the patterns and the verbalizers used by the {\PET} and \model models  for the SuperGLUE dataset here. For improved readability of the patterns, we first list a legend for the different letter combinations that we use throughout the patterns and then proceed to enumerate the patterns for each dataset.
\begin{itemize}
    \item \fancyp : passage/paragraph/pronoun
    \item \fancyq : question
    \item \fancyh : hypothesis
    \item \fancye : entity
    \item \fancyw : word
    \item \fancyci : choice i
    \item \fancysi : sentence i
\end{itemize}

\subsubsection{BoolQ \cite{clark2019boolq}} 
For this QA task, we are given a paragraph \fancyp and a yes/no question \fancyq. We use two forms of labels for this task yes/no and true/false.
\begin{itemize}
    \item \underline{Pattern} : \pattern{\fancyp. Question: \fancyq? Answer: \_\_\_.}\\
          \underline{Verbalizer}: \textbf{\fontfamily{lmtt}\selectfont yes/no}
    \item \underline{Pattern} : \\
    \begin{multipattern}
          \fancyp. Based on the previous passage, \fancyq? \_\_\_.
    \end{multipattern}
          \underline{Verbalizer}: \textbf{\fontfamily{lmtt}\selectfont yes/no}
    \item \underline{Pattern} : \\
    \begin{multipattern}
          Based on the following passage, \fancyq? \_\_\_. \fancyp
    \end{multipattern}
          \underline{Verbalizer}: \textbf{\fontfamily{lmtt}\selectfont yes/no}
    \item \underline{Pattern} : \pattern{\textbf{\fontfamily{lmtt}\selectfont p}. Question: \textbf{\fontfamily{lmtt}\selectfont q}? Answer: \_\_\_.}\\
          \underline{Verbalizer}: \textbf{\fontfamily{lmtt}\selectfont true/false}
    \item \underline{Pattern} : \\
    \begin{multipattern}
          \fancyp. Based on the previous passage, \fancyq? \_\_\_.
    \end{multipattern}
          \underline{Verbalizer}: \textbf{\fontfamily{lmtt}\selectfont true/false}
    \item \underline{Pattern} : \\
    \begin{multipattern}
         Based on the following passage, \fancyq? \_\_\_. \fancyp 
    \end{multipattern}
          \underline{Verbalizer}: \textbf{\fontfamily{lmtt}\selectfont true/false}
\end{itemize}

\subsubsection{CB \cite{de2019commitmentbank}}
In this textual entailment task, given a premise {\fancyp} and hypothesis {\fancyh} we need to determine if the {\fancyh} entails/contradicts/is neutral with respect to the {\fancyp}. The labels for this task are mapped to yes/no/maybe respectively. 
\begin{itemize}
    \item \underline{Pattern} : \pattern{\fancyh? $|$ \_\_\_,\fancyp}\\
          \underline{Verbalizer}: \textbf{\fontfamily{lmtt}\selectfont yes/maybe/no}
    \item \underline{Pattern} : \pattern{``\fancyh"? $|$ \_\_\_,``\fancyp"}\\
          \underline{Verbalizer}: \textbf{\fontfamily{lmtt}\selectfont yes/maybe/no}
    \item \underline{Pattern} : \pattern{\fancyh? $|$ \_\_\_.\fancyp}\\
          \underline{Verbalizer}: \textbf{\fontfamily{lmtt}\selectfont yes/maybe/no}
    \item \underline{Pattern} : \pattern{``\fancyh?" $|$ \_\_\_.``\fancyp"}\\
          \underline{Verbalizer}: \textbf{\fontfamily{lmtt}\selectfont yes/maybe/no}
\end{itemize}

\subsubsection{RTE \cite{dagan2005pascal}}
This is a textual entailment task similar to CB, except that we have just two labels for classification, entailment and not entailment. We map these two labels to yes and no respectively in the PVPs.
\begin{itemize}
    \item \underline{Pattern} : \pattern{\fancyh? $|$ \_\_\_,\fancyp}\\ \\
          \underline{Verbalizer}: \textbf{\fontfamily{lmtt}\selectfont yes/no}
    \item \underline{Pattern} : \pattern{``\fancyh"? $|$ \_\_\_,``\fancyp"} \\
          \underline{Verbalizer}: \textbf{\fontfamily{lmtt}\selectfont yes/no}
    \item \underline{Pattern} : \pattern{\fancyh? $|$ \_\_\_.\fancyp} \\ 
          \underline{Verbalizer}: \textbf{\fontfamily{lmtt}\selectfont yes/no}
    \item \underline{Pattern} : \pattern{``\fancyh?" $|$ \_\_\_.``\fancyp"} \\
          \underline{Verbalizer}: \textbf{\fontfamily{lmtt}\selectfont yes/no}
\end{itemize}

\subsubsection{COPA \cite{roemmele2011choice}}
Given a premise \fancyp, we need to find which of the options \textbf{\fontfamily{lmtt}\selectfont $\text{c}_1$} or \textbf{\fontfamily{lmtt}\selectfont $\text{c}_2$} is the responsible cause/effect for this task.
For effect examples:
\begin{itemize}
    \item \underline{Pattern} : \pattern{``\textbf{\fontfamily{lmtt}\selectfont $\text{c}_1$}" or ``\textbf{\fontfamily{lmtt}\selectfont $\text{c}_2$}”? \fancyp, so \_\_\_.}\\
          \underline{Verbalizer}: \textbf{\fontfamily{lmtt}\selectfont $\text{c}_1$/$\text{c}_2$}
    \item \underline{Pattern} : \pattern{\textbf{\fontfamily{lmtt}\selectfont $\text{c}_1$} or \textbf{\fontfamily{lmtt}\selectfont $\text{c}_2$}? \fancyp, so \_\_\_.}\\
          \underline{Verbalizer}: \textbf{\fontfamily{lmtt}\selectfont $\text{c}_1$/$\text{c}_2$}
\end{itemize}
For cause examples:
\begin{itemize}
    \item \underline{Pattern} : \pattern{``\textbf{\fontfamily{lmtt}\selectfont $\text{c}_1$}" or ``\textbf{\fontfamily{lmtt}\selectfont $\text{c}_2$}”? \fancyp, because \_\_\_.}\\
          \underline{Verbalizer}: \textbf{\fontfamily{lmtt}\selectfont $\text{c}_1$/$\text{c}_2$}
    \item \underline{Pattern} : \pattern{\textbf{\fontfamily{lmtt}\selectfont $\text{c}_1$} or \textbf{\fontfamily{lmtt}\selectfont $\text{c}_2$}? \fancyp, because \_\_\_.}\\
          \underline{Verbalizer}: \textbf{\fontfamily{lmtt}\selectfont $\text{c}_1$/$\text{c}_2$}
\end{itemize}

\subsubsection{WiC \cite{pilehvar2018wic}}
In this task, we are given two sentences \textbf{\fontfamily{lmtt}\selectfont $\text{s}_1$} and \textbf{\fontfamily{lmtt}\selectfont $\text{s}_2$} and we need to identify if a word {\fancyw} occurs in the same sense in both sentences.
\begin{itemize}
    \item \underline{Pattern} : \\
    \pattern{``\textbf{\fontfamily{lmtt}\selectfont $\text{s}_1$}" / ``\textbf{\fontfamily{lmtt}\selectfont $\text{s}_2$}" Similar sense of ``\fancyw"? \_\_\_ .}
          \underline{Verbalizer}: \textbf{\fontfamily{lmtt}\selectfont yes/no}
    \item \underline{Pattern} :\\
    \begin{multipattern}
          \textbf{\fontfamily{lmtt}\selectfont $\text{s}_1$} \textbf{\fontfamily{lmtt}\selectfont $\text{s}_2$} Does {\fancyw} have the same meaning in both sentences?\_.
    \end{multipattern}
          \underline{Verbalizer}: \textbf{\fontfamily{lmtt}\selectfont yes/no}
    \item \underline{Pattern} : \pattern{\fancyw. Sense (1) (a) ``\textbf{\fontfamily{lmtt}\selectfont $\text{s}_1$}" (\_ ) ``\textbf{\fontfamily{lmtt}\selectfont $\text{s}_2$}"}\\
          \underline{Verbalizer}: \textbf{\fontfamily{lmtt}\selectfont b/2}
\end{itemize}

\subsubsection{WSC \cite{levesque2012winograd}}
Here, we are given a sentence \textbf{\fontfamily{lmtt}\selectfont s} that contains some nouns and pronouns. We are tasked with finding the correct noun that a specific pronoun {\fancyp} refers to. Within the FewGLUE dataset, we are provided with the only positive examples and hence our verbalizer contains just the correct noun phrase.
\begin{itemize}
    \item \underline{Pattern} : \pattern{\textbf{\fontfamily{lmtt}\selectfont s} The pronoun `*\fancyp*' refers to \_\_\_.}\\
          \underline{Verbalizer}: \textbf{\fontfamily{lmtt}\selectfont correct noun}
    \item \underline{Pattern} :\\
    \begin{multipattern}
          \textbf{\fontfamily{lmtt}\selectfont s} In the previous sentence, the pronoun `*\fancyp*' refers to \_\_.
    \end{multipattern}
          \underline{Verbalizer}: \textbf{\fontfamily{lmtt}\selectfont correct noun}
    \item \underline{Pattern} :\\ 
    \begin{multipattern}
          \textbf{\fontfamily{lmtt}\selectfont s} In the passage above, what does the pronoun `*\fancyp*' refer to? Answer: \_\_.
    \end{multipattern}
          \underline{Verbalizer}: \textbf{\fontfamily{lmtt}\selectfont correct noun}
\end{itemize}

\subsubsection{MultiRC \cite{khashabi2018looking}}
In this task, we are given a passage \fancyp and multiple questions \fancyq. We are tasked with finding the right answer from a list of candidate answers \fancye. Here, we pose it as a binary classification task where we predict yes if the {\fancye} answers {\fancyq} with context {\fancyp}, else no.
\begin{itemize}
    \item \underline{Pattern} : \pattern{\fancyp. Question: \fancyq ? Is it \fancye? \_\_\_.}\\
          \underline{Verbalizer}: \textbf{\fontfamily{lmtt}\selectfont yes/no}
    \item \underline{Pattern} : \\
    \begin{multipattern}
          \fancyp. Question: \fancyq? Is the correct answer ``\fancye"? \_\_\_.
    \end{multipattern}
          \underline{Verbalizer}: \textbf{\fontfamily{lmtt}\selectfont yes/no}
    \item \underline{Pattern} : \\ 
    \begin{multipattern}
          {\fancyp}. Based on the previous passage, \fancyq? Is ``\fancye" a correct answer? \_\_.
    \end{multipattern}
          \underline{Verbalizer}: \textbf{\fontfamily{lmtt}\selectfont yes/no}
\end{itemize}

\subsubsection{ReCoRD \cite{zhang2018record}}
For this task, given a passage {\fancyp} and cloze question \fancyq, we are supposed to find the right replacement for a `\texttt{@placeholder}' token in the question. Since the task itself is already framed in a cloze-style format, we merely concatenate the passage with the cloze question to form the input to the language model.

\subsection{Results on Individual Patterns} \label{sec:ind_pvp}

We train the s{\PET} and \model models using the same experimental setup mentioned in Section \ref{sec:exp} and report results across all patterns for all datasets on the validation dataset of SuperGLUE. Note that the numbers in Table \ref{tab:res} contains the best numbers from this table for the dev results. Our results can be found in Table \ref{tab:pattern}. Overall, \model outperforms s{\PET} on 25 out of 29 patterns across datasets.

\begin{table*}
    \resizebox{\textwidth}{!}{
    \centering
    \begin{tabular}{c| c | l l l l l l }
        \hline
            & \textbf{\begin{tabular}[c]{@{}c@{}}Pattern/\\ Model\end{tabular}} & 1 & 2 & 3 & 4 & 5 & 6 \\ 
        \hline
        
        \multirow{2}{*}{\begin{tabular}[c]{@{}c@{}}\textbf{BoolQ}\\ Acc.\end{tabular}} & s{\PET} & 75.8 & 76.9 $\reddiamond$ & 74.6 & 76.0 & 76.3 & 68.4\\ 
          & \model   & 80.3 $\clubsuit \reddiamond$ & 78.6  $\clubsuit$ & 80.0  $\clubsuit$ & 78.1  $\clubsuit$ & 78.0  $\clubsuit$ & 79.9  $\clubsuit$\\
        \hline
        
        \multirow{2}{*}{\begin{tabular}[c]{@{}c@{}}\textbf{CB}\\ Acc./F1\end{tabular}} & s{\PET} & 75/72.8 & 87.5/85.4 $\reddiamond$  & 83.9/68.9 & 85.7/82.3 & - & - \\ 
        & \model & 89.3/81.4  $\clubsuit$ & 89.3/86.8  $\clubsuit \reddiamond$ & 89.3/85.2  $\clubsuit$ & 89.3/86.8  $\clubsuit$& - & - \\
        \hline
        
        \multirow{2}{*}{\begin{tabular}[c]{@{}c@{}}\textbf{COPA}\\ Acc.\end{tabular}} & s{\PET} & 89 $\reddiamond$ & 85 $\clubsuit$ & - & - & - & - \\ 
        & \model & 89 $\reddiamond$& 77 & - & - & - & - \\
        \hline
        
        \multirow{2}{*}{\begin{tabular}[c]{@{}c@{}}\textbf{MultiRC}\\ F1a/EM\end{tabular}} & s{\PET} & 30.6/73.7 & 29.9/73.2 & 19.1/65 & 30.8/74.6 $\reddiamond$ & 15/65.2 & 23.1/69.6 \\ 
        & \model & 35.8/79.1 $\clubsuit$ & 34.7/78.3 $\clubsuit$ & 39.2/80.1 $\clubsuit \reddiamond$ & 35.7/78.2 $\clubsuit$ & 35.5/78.9 $\clubsuit$ & 31.5/76.8 $\clubsuit$ \\ 
        \hline
        
        \multirow{2}{*}{\begin{tabular}[c]{@{}c@{}}\textbf{RTE}\\ Acc.\end{tabular}} & s{\PET} & 56 & 53.8 & 59.9 & 67.1 $\reddiamond$& - & - \\ 
        & \model & 76.2 $\clubsuit$ & 75.1 $\clubsuit$ & 74.4 $\clubsuit$ & 76.5 $\clubsuit \reddiamond$ & - & -\\
        \hline
        
        \multirow{2}{*}{\begin{tabular}[c]{@{}c@{}}\textbf{WiC}\\ Acc.\end{tabular}} & s{\PET} & 49.7 $\clubsuit \reddiamond$ & 47.5 & 49.7 & - & - & - \\ 
        & \model & 49.4 & 52.4 $\clubsuit$ & 54.5 $\clubsuit \reddiamond$ & - & - & - \\
        \hline
        
        \multirow{2}{*}{\begin{tabular}[c]{@{}c@{}}\textbf{WSC}\\ Acc.\end{tabular}} & s{\PET} & 82.7 $\clubsuit \reddiamond$ & 79.8 & 81.7 $\clubsuit$ & - & - & - \\ 
        & \model & 81.7 $\reddiamond$& 79.8 & 79.8 & - & - & - \\ \hline
        
        \multirow{2}{*}{\begin{tabular}[c]{@{}c@{}}\textbf{ReCoRD}\\ Acc./F1\end{tabular}} & s{\PET} & 85.0/91.9 $\reddiamond$ & - & - & - & - & - \\ 
        & \model & 85.4/92.1 $\clubsuit \reddiamond$ & - & - & - & - & - \\
        \hline
    \end{tabular}
    }
    \caption{Performance of s{\PET} and \model models on the validation set of SuperGLUE for different patterns after training for 1000 batches. The patterns we use are the same as {\PET} \cite{schick2020s}. Note that Table \ref{tab:res} uses the best pattern ($\reddiamond$) results from this table for each model to report validation set scores.\textsc{$\clubsuit$ = Best Model for Each Pattern}}
    \label{tab:pattern}
\end{table*}

\begin{table*}
    \resizebox{\textwidth}{!}{
    \centering
    \begin{tabular}{c| c | l l l l l l }
        \hline
            & \textbf{\begin{tabular}[c]{@{}c@{}}Pattern/\\ Model\end{tabular}} & 1 & 2 & 3 & 4 & 5 & 6 \\ 
        \hline
        
        \multirow{2}{*}{\begin{tabular}[c]{@{}c@{}}\textbf{BoolQ}\\ Acc.\end{tabular}} & s{\PET} & 77.5  & 77.1 & 73.9 & 75.6 & 74.2 & 66.8 \\ 
          & \model   & 79.4 $\clubsuit$ & 78.3 $\clubsuit$ & 78.7 $\clubsuit$ & 77.7 $\clubsuit$& 78.2 $\clubsuit$ & 76.8 $\clubsuit$\\
        \hline
        
        \multirow{2}{*}{\begin{tabular}[c]{@{}c@{}}\textbf{CB}\\ Acc./F1\end{tabular}} & s{\PET} & 75/72.8 & 85.7/83.5   & 83.9/68.9 & 85.7/82.3 & - & - \\ 
        & \model & 91.1/88.1 $\clubsuit$ & 87.5/85.5 $\clubsuit$ & 87.5/78.7 $\clubsuit$ & 89.3/85 $\clubsuit$ & - & - \\
        \hline
        
        \multirow{2}{*}{\begin{tabular}[c]{@{}c@{}}\textbf{COPA}\\ Acc.\end{tabular}} & s{\PET} & 90 $\clubsuit$ & 87 & - & - & - & - \\ 
        & \model & 73 & 89 $\clubsuit$ & - & - & - & - \\
        \hline
        
        \multirow{2}{*}{\begin{tabular}[c]{@{}c@{}}\textbf{MultiRC}\\ F1a/EM\end{tabular}} & s{\PET} & 29.9/72.8 & 30.2/73.3  & 23.6/69.0 & 27.4/72.8  & 16.1/65.7 & 23.9/70.3 \\ 
        & \model & 36.4/79.4 $\clubsuit$ & 36.0/78.6 $\clubsuit$ & 38.1/79.0 $\clubsuit$ & 34.6/77.9 $\clubsuit$ & 33.2/77.8 $\clubsuit$ & 31.4/75.1 $\clubsuit$\\ 
        \hline
        
        \multirow{2}{*}{\begin{tabular}[c]{@{}c@{}}\textbf{RTE}\\ Acc.\end{tabular}} & s{\PET} & 57 & 54.5 & 56.7 & 71.7 & - & - \\ 
        & \model & 74.7$\clubsuit$  & 69.7 $\clubsuit$ & 75.1 $\clubsuit$ & 73.6 $\clubsuit$ & - & -\\
        \hline
        
        \multirow{2}{*}{\begin{tabular}[c]{@{}c@{}}\textbf{WiC}\\ Acc.\end{tabular}} & s{\PET} & 49.8 & 47.8 & 49.5 & - & - & - \\ 
        & \model & 51.1 $\clubsuit$ & 49.5 $\clubsuit$ & 50.8 $\clubsuit$ & - & - & - \\
        \hline
        
        \multirow{2}{*}{\begin{tabular}[c]{@{}c@{}}\textbf{WSC}\\ Acc.\end{tabular}} & s{\PET} & 82.7 $\clubsuit$ & 78.8 $\clubsuit$ & 79.8 & - & - & - \\ 
        & \model & 76.9 & 74 & 79.8 & - & - & - \\ \hline
        
        \multirow{2}{*}{\begin{tabular}[c]{@{}c@{}}\textbf{ReCoRD}\\ Acc./F1\end{tabular}} & s{\PET} & 82.3/91 $\clubsuit$ & - & - & - & - & - \\ 
        & \model & 77.4/87.2 & - & - & - & - & - \\
        \hline
    \end{tabular}
    }
    \caption{Performance of s{\PET} and \model models on the validation set of SuperGLUE for different patterns after training for 250 batches. The patterns we use are the same as {\PET} \cite{schick2020s}. \textsc{$\clubsuit$ = Best Model for Each Pattern}}
    \label{tab:pattern_250}
\end{table*}

\section{(More) Experiment Details} \label{sec:exp_det}

\subsection{Decoupled Label Objective}

All our experiments followed the same setup as {\PET} \cite{schick2020exploiting}. We use a random seed of $42$, maximum text length of $256$ \footnote{Note: for MultiRC and ReCoRD we use 512 tokens as per \cite{schick2020s}.}, AdamW optimizer, learning rate of $1e^{-5}$, weight decay of $1e^{-2}$, and linear decay scheduler with a warmup over the first $10\%$ of batches.

\subsection{Label Conditioning Objective}

For all datasets, we mask out up to $10.5\%$ of tokens in the text. For COPA, because the pattern contains both the correct and incorrect choice, we use a different pattern where we only feed in one choice for the label conditioning objective.

For the cause examples:
\begin{itemize}
    \item \underline{Pattern} : 
    \pattern{Because \fancyp , \_\_\_.}\\
          \underline{Verbalizer}: \textbf{\fontfamily{lmtt}\selectfont $\text{c}_1$/$\text{c}_2$}
\end{itemize}

For the effect examples:
\begin{itemize}
    \item \underline{Pattern} : 
    \pattern{Because \_\_\_ , \fancyp.}\\
          \underline{Verbalizer}: \textbf{\fontfamily{lmtt}\selectfont $\text{c}_1$/$\text{c}_2$}
\end{itemize}

\section{Ablations} \label{sec:ablat}

\subsection{Duration of Training}

We trained \model for 1k batches and compared to {\PET}/i{\PET} which were trained for 250 batches. In this section, we compare s{\PET} and \model trained for 250 and 1k batches in Table \ref{tab:dur_train}. Note that training for 1k batches is not guaranteed to outperform training for 250 batches, even if we checkpoint every 250 batches, since the learning rate scheduler will have to accommodate for a different number of total batches. Overall, \model gets a boost by training longer, especially on ReCoRD, while s{\PET} peaks at 250 batches.

\begin{table*}[ht!]
    \centering
    \resizebox{1\textwidth}{!}{%
    \begin{tabular}{l l l l l l l l l l | l}
        \hline 
        & & \textbf{BoolQ} & \textbf{CB} & \textbf{COPA} & \textbf{RTE} & \textbf{WiC} & \textbf{WSC} & \textbf{MultiRC} & \textbf{ReCoRD} & \textbf{Avg} \\
        & \textbf{Method} & Acc. & Acc./F1 & Acc. & Acc. & Acc. & Acc. & EM/F1a & Acc./F1 & -\\
        \hline
        \multirow{4}{*}{\rotatebox[origin=c]{90}{dev}}   
        & s{\PET} \textsc{(lab; single; 250 Batches)}  & \textbf{77.5} &  85.7 / 83.5 & \textbf{90.0} & \textbf{71.7} & \textbf{49.8} & 82.7  & 30.2 / 73.3 & 82.3 / 91.0 & \textbf{74.3} \\
         & s{\PET} \textsc{(lab; single; 1k Batches)}  & 76.9 &  \textbf{87.5 / 85.4} & 89.0 & 67.1 & 49.7 & 82.7 & \textbf{31.2 / 74.6} & \textbf{85.0 / 91.9} & 74.2 \\
        \cline{2-11}
        & \model \textsc{(lab; single; 250 Batches)}   & 79.4  & \textbf{91.1 / 88.1}  & 89.0 & 75.1  & 51.1 & 79.8 & 38.1 / 79.0  & 77.4 / 87.2  & 75.6 \\
        & \model \textsc{(lab; single; 1k Batches)}   & \textbf{80.3}  & 89.3 / 86.8 & 89.0 & \textbf{76.5}  & \textbf{54.4} & \textbf{81.7} & \textbf{39.2 / 80.1}  & \textbf{85.4 / 92.1}  &  \textbf{77.3}  \\
        \hline
        \multirow{2}{*}{\rotatebox[origin=c]{90}{test}}   
        & \model \textsc{(lab; single; 250 Batches)}  & 78.4  & \textbf{93.6 / 86.4}  & \textbf{86.0} & \textbf{75.0}  & 49.6 &  \textbf{90.4}  & \textbf{37.3 / 75.4}  & 78.5 / 79.5 & 75.6 \\
        & \model \textsc{(lab; single; 1k Batches)}  & \textbf{80.0} & 92.0 / 82.3  & 85.4 & \textbf{75.0} & \textbf{53.5} &  85.6  & 35.7 / 76.2 & \textbf{85.5 / 86.1} & \textbf{76.0} \\
        \hline
    \end{tabular}
    }
    \caption{Performance of the models trained with 250 batches vs 1k batches }
    \label{tab:dur_train}
\end{table*}

\subsection{Multi-Task Multi-Pattern Training}

We also tried training the model with multiple patterns at once, as compared to ensembling and distilling them. We formulated this as a multi-task training problem, where different patterns are viewed as different tasks, and the model would sample a pattern to train from each batch. We compare s{\PET}, \model, and \model without the label conditioning objective. The results are shown in Table \ref{tab:mp_mt}. In general, multi-task multi-pattern training hurts performance for \model, is mixed on s{\PET}, and is beneficial for \model with the label conditioning objective. 

\begin{table}[ht!]
    \centering
    \resizebox{0.5\textwidth}{!}{%
    \begin{tabular}{l c c c c}
        \hline 
        &  \textbf{BoolQ} & \textbf{CB} & \textbf{RTE} & \textbf{MultiRC}\\
        \textbf{Method} &  Acc. & Acc./F1 &  Acc. & EM / F1a\\
         \hline
         s{\PET} & \textbf{77.5} &	\textbf{85.7/83.5} & 71.7 & 30.2 / 73.3 \\
         s{\PET} \textsc{(mtmp)} & 77.3 &	87.5/78.7  &  \textbf{74} & \textbf{30.1 / 74.3}\\
        \hline
        \model  &  \textbf{79.4} & \textbf{91.1 / 88.1}	& \textbf{75.1} & \textbf{38.1 / 79.0} \\
        \model \textsc{(mtmp)} &  78.9	& 89.3/86.8	& 73.3 & 35.9/78.3 \\
        \hline
        \model { \textsc{w/o LC}}  & 77.8 & 78.6 / 54.9  & 71.5 & \textbf{32.5 / 74.8}  \\
        \model \textsc{\textsc{w/o LC} (mtmp)} &  \textbf{79.9} & \textbf{89.3/83.6} & \textbf{77.3} & 27.7/72.6 \\

        \hline
    \end{tabular}
    }
    \caption{Comparison of s{\PET} and \model with Multi-Pattern Multi-Task training \textsc{mpmt = Multi Pattern Multi Task}. Best numbers have been \textbf{bolded}. (\textsc{LC= Label Conditioning}) }
    \label{tab:mp_mt}
\end{table}

\subsection{Replacement Token Detection (RTD)}

In our formulation, the decoupled label objective can be viewed as a binary classifier that seeks to assign high probability to the correct label token, and low probability to the incorrect label token. In reality though, the model has a softmax classifier head on top that is converted into a one-vs-all classifier. 

Another way to achieve the same objective would be to use a binary classifier head on top. Rather than feeding in the \textit{``[MASK]"} token, we would feed in either the correct label token or the incorrect label token, and the model must distinguish whether these tokens make sense in context or not. This objective would be very similar to the RTD objective for ELECTRA \cite{clark2020electra}. 
Inference would be slower since the number of forward passes would scale up by the number of labels. For multi token labels though, because there is not need to condition on other label tokens, the number of forward passes would scale down by the number of tokens in the labels.  

Table \ref{tab:real_disc} shows the results of using the RTD objective with a binary classifier. Overall, the RTD objective seems to perform worse than the decoupled label objective. There are several reasons why using a RTD head might perform worse. First, the RTD head would have $|V|$ times fewer parameters, but relative to the whole model, the change in number of parameters is not substantial. Second, the softmax classifier has been pretrained, and contains lots of information, which is now lost when we discard the softmax classifier and randomly initialize a binary classifier head from scratch. 

We also experiment with using a binary classifier head initialized with ELECTRA, but the results were the same and so we omit them from the table. We note that ALBERT (xxlarge-v2) is a much better performing model than BERT, and ELECTRA is more comparable to BERT than  ALBERT (xxlarge-v2).

\begin{table}[ht!]
    \centering
    \resizebox{0.5\textwidth}{!}{%
    \begin{tabular}{l c c c c c}
        \hline 
        &  \textbf{BoolQ} & \textbf{CB} & \textbf{RTE} & \textbf{MultiRC} \\
        \textbf{Method} &  Acc. & Acc./F1 &  Acc. & EM / F1a\\
         \hline
        \model \small{\textsc{w/o LC}}  &  \textbf{77.8} & 78.6 / 54.9  & \textbf{71.5} & \textbf{32.5 / 74.8} \\
        \model \small{RTD}  &  69.8 &	\textbf{82.1 / 80.2}  &  57.8 & 21.7 / 72.2  \\
        \hline
    \end{tabular}
    }
    \caption{Comparison of decoupled label objective and with the replacement token detection (RTD) objective. Best numbers have been \textbf{bolded}.  (\textsc{LC= Label Conditioning})}
    \label{tab:real_disc}
\end{table}

\subsection{Label Conditioning with Important Words Masked Out}

For the label conditioning component, we randomly mask out tokens in the input text, and the model tries to predict the original token when conditioned on the correct label, and not predict the original token when conditioned on an incorrect label. This makes sense if the masked out token is an influential token that affects the label, like \textit{``Yes"} in Figure \ref{fig:decouple}, but makes less sense if the masked out token is an unimportant word like \textit{``the"}. We experiment with only masking out important words, using TFIDF as an approximation of how important a word is. The results are shown in table \ref{tab:tfidf}. Overall, using TFIDF as an approximation for masking out important words hurts performance.  

\begin{table}[ht!]
    \centering
    \resizebox{0.5\textwidth}{!}{%
    \begin{tabular}{l c c c c c}
        \hline 
        &  \textbf{BoolQ} & \textbf{CB} & \textbf{RTE} & \textbf{MultiRC} \\
        \textbf{Method} &  Acc. & Acc./F1 &  Acc. & EM / F1a\\
         \hline
        \model  &  \textbf{79.4}	& \textbf{91.1 / 88.1}	& \textbf{74.7} & \textbf{36.4 / 79.4} \\
        \model \small{TFIDF}  &  76.1 &	76.8/61.8  &  72.9 & 31.1 / 77.1  \\
        \hline
    \end{tabular}
    }
    \caption{Comparison of \model with random masking and masking tokens based on TFIDF. Best numbers have been \textbf{bolded}.  (\textsc{LC= Label Conditioning})}
    \label{tab:tfidf}
\end{table}

\subsection{Ensembles}

{\PET}/i{\PET} ensemble and distill with unlabeled data. However, it is not clear how beneficial unlabeled data is for ensembling, so we show results of ensembling models trained only on labeled data with different patterns and different seeds. For ensembling, we average the logits across the different models. 

\subsubsection{Across Patterns}

Table \ref{tab:ens_pat} shows our results ensembling across patterns. In general, ensembling across patterns provides mixed results for \model and s{\PET}. This corroborates the finding in \citet{gao2020making} where sometimes the best performing model performs better than ensembling across patterns. 

\subsubsection{Across Seeds}

Table \ref{tab:ens_seed} shows our results ensembling across seeds. We fix the pattern (pattern 1) and train with different seeds. For this experiment, we ensemble across models for seeds 41, 42, 43. From our results in Table \ref{tab:ens_seed}, we find that ensembling patterns across seeds provides mixed results. Hence, we do not apply ensembling for our final results.

\begin{table}[ht!]
    \centering
    \resizebox{0.5\textwidth}{!}{%
    \begin{tabular}{l c c c c c}
        \hline 
        &  \textbf{BoolQ} & \textbf{CB} & \textbf{RTE} & \textbf{MultiRC} \\
        \textbf{Method} &  Acc. & Acc./F1 &  Acc. & EM / F1a\\
         \hline
        \model  &  79.4	& \textbf{91.1 / 88.1}	& \textbf{75.1} & 38.1/ 79.0 \\
        \model \textsc{(ens; pat)}  & \textbf{79.5}	& 89.3/86.8	& \textbf{75.1} & \textbf{38.2/79.2} \\
        \hline
         s{\PET} & 77.5 & \textbf{85.7/83.5} & 71.7 & 30.2 / 73.3 \\
         s{\PET} \textsc{(ens; pat)}  & \textbf{78.2} & 71.4 / 77.8 & \textbf{74.3} & \textbf{30.7 / 73.8} \\ 
         \hline
    \end{tabular}
    }
    \caption{Ensemble of s{\PET} and \model across patterns. We use the best pattern (instead of pattern 1) numbers for \model and s{\PET} here. (\textsc{ens= Ensemble}) (\textsc{pat= Pattern}) Best numbers have been \textbf{bolded}.}
    \label{tab:ens_pat}
\end{table}

\begin{table}[ht!]
    \centering
    \resizebox{0.5\textwidth}{!}{%
    \begin{tabular}{l c c c c c}
        \hline 
        &  \textbf{BoolQ} & \textbf{CB} & \textbf{RTE} & \textbf{MultiRC} \\
        \textbf{Method} &  Acc. & Acc./F1 &  Acc. & EM / F1a\\
         \hline
        \model  &  \textbf{79.4}	& \textbf{91.1 / 88.1}	& \textbf{75.1} & \textbf{38.1/ 79.0} \\
        \model \textsc{(ens; seed)}  & 79 & \textbf{91.1 / 88.1} & 69 & 35.9 / 79.3 \\
        \hline
         s{\PET}  & 77.5&  \textbf{75.0 / 72.8} & \textbf{57.0} & 26.5 / 73.2\\
         s{\PET} \textsc{(ens; seed)}  &  \textbf{77.8} & 78.6 / 64.1 & 53.1 & \textbf{30.5 / 73.8} \\
         \hline
    \end{tabular}
    }
    \caption{Ensemble of s{\PET} and \model across seeds. Best numbers have been \textbf{bolded}.}
    \label{tab:ens_seed}
\end{table}

\subsection{Masking Ratio}

We experiment with several different masking schemes, where we mask out a fixed percentage (\textsc{Fixed}), or \textit{up to} a fixed percentage (\textsc{Variable}) in Table \ref{tab:mask_ratio}. If $x$ is the number of tokens masked out in \textsc{Fixed} masking, we mask out between $1$ and $x$ tokens for \textsc{Variable} masking. For the ablation, we tested with multiples of $1.5$ for the masking ratio (in addition to $10\%$), to match the $15\%$ ratio of ALBERT pre-training. From our results in Table \ref{tab:mask_ratio}, we find that $10.5\%$ \textsc{Variable} mask ratio provided the best trade-off between scores for all models. Hence, we choose that for our final experiments in the main paper.

\begin{table}[ht!]
    \centering
    \resizebox{0.5\textwidth}{!}{%
    \begin{tabular}{l c c c c c}
        \hline 
        &  \textbf{BoolQ} & \textbf{CB} & \textbf{RTE} & \textbf{MultiRC} \\
        \textbf{Masking Ratio} &  Acc. & Acc./F1 &  Acc. & EM / F1a\\
         \hline
        15\% \textsc{(fixed)} & 80.7 & 91.1/87.7 & 70.8 & 35.8/79.1 \\
        10.5\% \textsc{(fixed)} & 80.1 & 89.3/85.0 & 72.9 & 35.8/79.1 \\
        10\% \textsc{(fixed)} & 79.9 & 81.1/87.5 & 69.0 & 33.9/78.4 \\
        7.5\% \textsc{(fixed)} & 78.3 & 85.7/79.8 & 74 & 36.9/78.8 \\
        \hline
        15\% \textsc{(variable)}& 78.9 & 87.5/80.0 & 75.1 & 35.9/78.7 \\
        10.5\% \textsc{(variable)} & 79.4 & 91.1/88.1 & 74.7 & 36.4/79.4 \\
        10\% \textsc{(variable)} & 80.0 & 89.3/86.8 & 71.1 & 33.9/78.4 \\
        7.5\% \textsc{(variable)} & 79.7 &  89.3/86.8  & 70.8 & 36.9/78.8\\
         \hline
    \end{tabular}
    }
    \caption{Results with different masking strategies for label-conditioned MLM in \model.}
    \label{tab:mask_ratio}
\end{table}

\subsection{What if we had unlabeled data?}

One of the key motivations of our work is to eliminate the need for unlabeled data during few-shot training on language understanding tasks. In this section, we push that limitation of prior methods aside and seek to know ``if" such unlabeled data were available, can \model leverage unlabeled data to improve performance. Instead of adopting the multi-stage iterative approach in iPET, we experiment with  pre-training the model on the unlabeled data before fine-tuning on the labeled dataset. This has been shown to improve performance on text-classification tasks previously \cite{gururangan-etal-2020-dont}. Specifically, we experiment with Task Adaptive Pre-training (TAPT)  \cite{gururangan-etal-2020-dont} and pre-train our base LM for 2500 batches on the unlabeled data of FewGLUE. Following that, we fine-tune the models using \model, s{\PET} and regular (CLS-head) fine-tuning on the labeled set. The results can be found in Table \ref{tab:taft}. For regular fine-tuning, TAPT improves performance on three out of four datasets. However, for s{\PET} and \model, TAPT hurts performance significantly for all datasets. We speculate this is because during TAPT, the model never sees the pattern, and so it hurts pattern-based models. This leaves the question of how to improve pattern-based few-shot methods, like \model, when unlabeled data is available as an open challenge. 

\begin{table}[ht!]
    \centering
    \resizebox{0.5\textwidth}{!}{%
    \begin{tabular}{l c c c c c}
        \hline 
        &  \textbf{BoolQ} & \textbf{CB} & \textbf{RTE} & \textbf{MultiRC} \\
        \textbf{Method} &  Acc. & Acc./F1 &  Acc. & EM / F1a\\
         \hline
        \model  &  \textbf{79.4}	& \textbf{91.1 / 88.1}	& \textbf{74.7} & \textbf{36.4 / 79.4} \\
        TAPT + \model  &  60.9 & 66.1/46.1 & 47.7 & 9.3 / 60.3 \\
        \hline
        s{\PET}  &  \textbf{77.5}	& \textbf{85.7 / 83.5}	& \textbf{71.7} & \textbf{30.2 / 77.3} \\
        TAPT + s{\PET}  &  62.9 & 69.6/58.9 & 44.8 & 5.5 / 58.1 \\
        \hline
        ALBERT  &   55.7 &  68.6/49.1	&  \textbf{50.5} &  3.6/49.8 \\
        TAPT + ALBERT  &  \textbf{60.6} & \textbf{69.6/58.9} & 47.7 & \textbf{6.3/54.1} \\
        \hline
    \end{tabular}
    }
    \caption{Results of TAPT pre-training with various models.  Best numbers have been \textbf{bolded}.}
    \label{tab:taft}
\end{table}

\end{document}